\documentclass[runningheads]{llncs}
\usepackage{graphicx}
\usepackage{tikz}
\usepackage{comment} 
\usepackage{amsmath,amssymb} 
\usepackage{color}
\usepackage{epsfig}
\usepackage{float}
\usepackage{subfigure}
\usepackage{enumerate}
\usepackage{inline-images}
\usepackage{setspace}
\usepackage{booktabs}
\usepackage{tabularx}
\usepackage{array}
\usepackage{blindtext}
\usepackage{caption}
\usepackage{multirow}
\usepackage{amssymb}
\usepackage{multirow}
\usepackage[misc]{ifsym}
\usepackage{bbding}
\usepackage[pagebackref=true,breaklinks=true,letterpaper=true,colorlinks,bookmarks=false]{hyperref}

\begin{document}
\pagestyle{headings}
\mainmatter
\def\ECCVSubNumber{5989}  

\title{Toward Fine-grained Facial Expression Manipulation}

\titlerunning{Toward Fine-grained Facial Expression Manipulation}
%
\author{Jun Ling\inst{1}\and
Han Xue\inst{1} \and
Li Song\inst{1,2}\inst{(}\Envelope\inst{)} \and
Shuhui Yang\inst{1}\and
Rong Xie\inst{1} \and
Xiao Gu\inst{1}}
\authorrunning{J. Ling et al.}

\institute{Institute of Image Communication and Network Engineering, Shanghai Jiao Tong University
\and
MoE Key Lab of Artificial Intelligence, AI Institute, Shanghai Jiao Tong University
\email{\{lingjun,xue\_han,song\_li,louisxiii,xierong,gugu97\}@sjtu.edu.cn}}
\maketitle

\begin{figure}
\centering
   \includegraphics[width=1\linewidth]{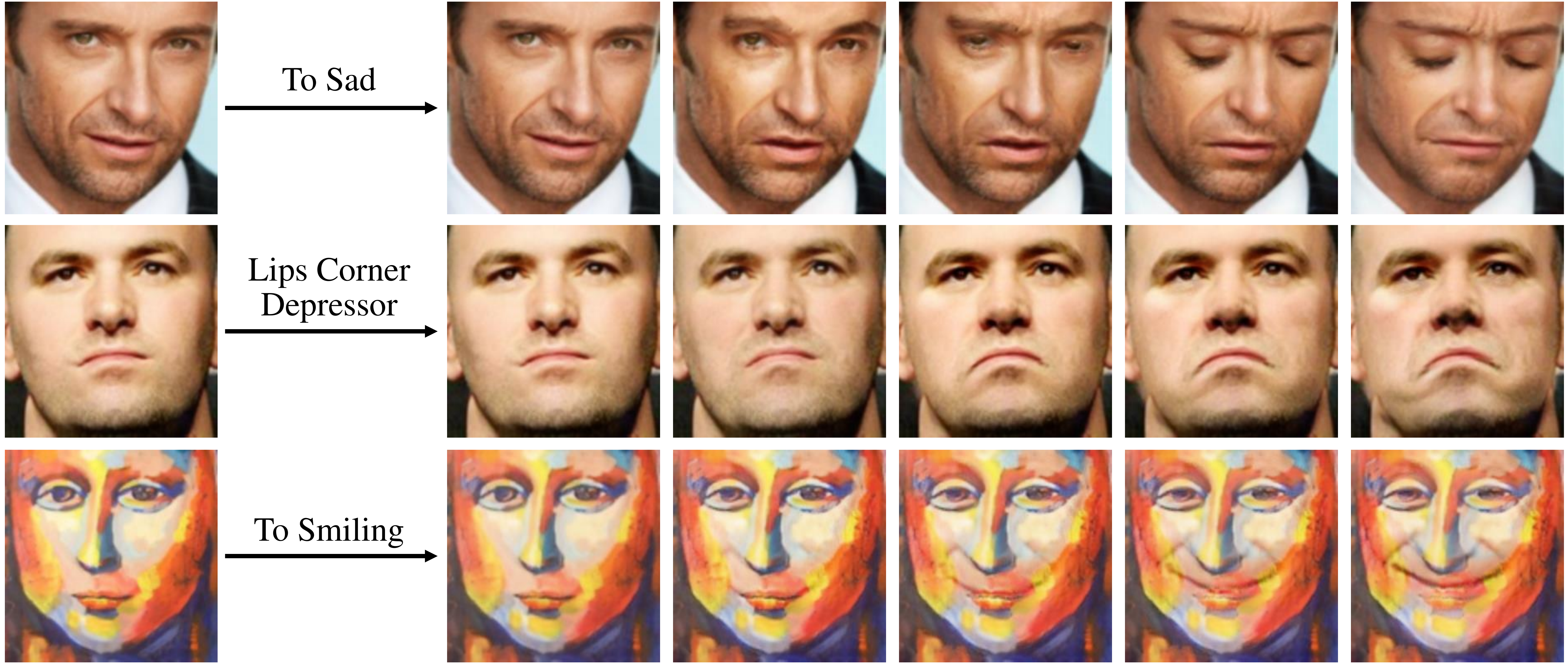}\\
   \caption{\textbf{Arbitrary Facial Expression Manipulation.} Our model can 1) perform continuous editing between two expressions (\emph{top}); 2) learn to only modify one facial component(\emph{middle}); 3) transform expression in paintings (\emph{bottom}). From left to right, the emotion intensity is set to 0, 0.5, 0.75, 1, and 1.25. }
\label{fig:expression_editing_intro}
\end{figure}

\begin{abstract}
Facial expression manipulation aims at editing facial expression with a given condition. Previous methods edit an input image under the guidance of a discrete emotion label or absolute condition (e.g., facial action units) to possess the desired expression. However, these methods either suffer from changing condition-irrelevant regions or are inefficient for fine-grained editing. In this study, we take these two objectives into consideration and propose a novel method. First, we replace continuous absolute condition with relative condition, specifically, relative action units. With relative action units, the generator learns to only transform regions of interest which are specified by non-zero-valued relative AUs. Second, our generator is built on U-Net but strengthened by multi-scale feature fusion (MSF) mechanism for high-quality expression editing purposes. Extensive experiments on both quantitative and qualitative evaluation demonstrate the improvements of our proposed approach compared to the state-of-the-art expression editing methods. Code is available at \url{https://github.com/junleen/Expression-manipulator}.
\end{abstract}

\section{Introduction}
Over the years, facial expression synthesis has been drawing considerable attention in the field of both computer vision and computer graphics. However, synthesizing easy-to-use and fine-grained facial images with desired expression remains challenging because of the complexity of this task. Recently, the proposal of generative adversarial networks~\cite{Goodfellow2014Generative,Mirza2014Conditional} sheds light on image synthesis, introducing significant advances with well-known architectures like~\cite{Choi_2018_CVPR_stargan,he2019attgan,Zhang2018sagenerative,liu2019stgan}. However, these work suffer from fine-grained expression editing because they either rely on several binary emotion labels (e.g., smiling, mouth open) to synthesize target expressions, or suffer from limited naturalness and low quality.

As one of the most successful generative models, GANimation~\cite{Pumarola_ijcv2019} pushes the limits of facial expression manipulation by building a conditional GAN which relies on attention-based generator and discrete facial action units activation (action units~\cite{friesen1978facial}(AUs), a kind embedding which indicates the facial muscles movement). As a novel expression editing method, GANimation is able to edit an image in a continuous manner and outperforms other popular multi-domain image-to-image translation methods~\cite{li2016diat,perarnau2016invertible,zhu2017cyclegan,Choi_2018_CVPR_stargan}. 

Despite the novelty and generality, GANimation suffers from two drawbacks. First, by taking absolute AUs as input condition, the generator needs to estimate the current facial muscles state so that it can apply a desired expression change to the input image. This is insufficient for the model to reserve its facial part corresponding to unchanged AUs. Besides, from the perspective of model testing, exploiting the entire set of AUs as conditional input imposes a restriction on fine-grained expression editing because a user always needs to acquire accurate underlying real value of each AU in the input image, even though he does not intend to to modify these facial regions. Second, the attention mechanism which is introduced for learning desirable change from expression of input image to desired expression, virtually applies a learned weighted sum between the input image and the generated one. This kind of operation, as pointed out in~\cite{Pumarola_ijcv2019}, brings about overlap artifacts around face deformation regions. Furthermore, spatial attention networks for attribute-specific region editing~\cite{Zhang2018sagenerative} are effective only for local attributes and not designed for arbitrary attribute editing~\cite{liu2019stgan}.

To address these limitations, this work investigates arbitrary facial expression editing on the basis of relative condition. In terms of \emph{relative}, which is defined as the difference between target AUs and source AUs, our model is capable of (i) only considering the facial components to be modified while keeping the remaining parts unchanged, and (ii) freely strengthening or suppressing the intensity of specified AUs or arbitrary emotions by user-input real numbers. This brings several benefits. First, by using relative AUs, the generator is not required to compare the current AUs with desired AUs before applying image transformation. Second, the values of the relative AUs indicate the desired change to facial muscles. In particular, non-zero values correspond to AUs of interest and zero values correspond to unchanged AUs. Hence, our generator can learn to manipulate single AU with scalable one-hot vector, eliminating the demand for all other AUs intensities. 

\begin{figure}[!t]
\centering
   \includegraphics[width=1\linewidth]{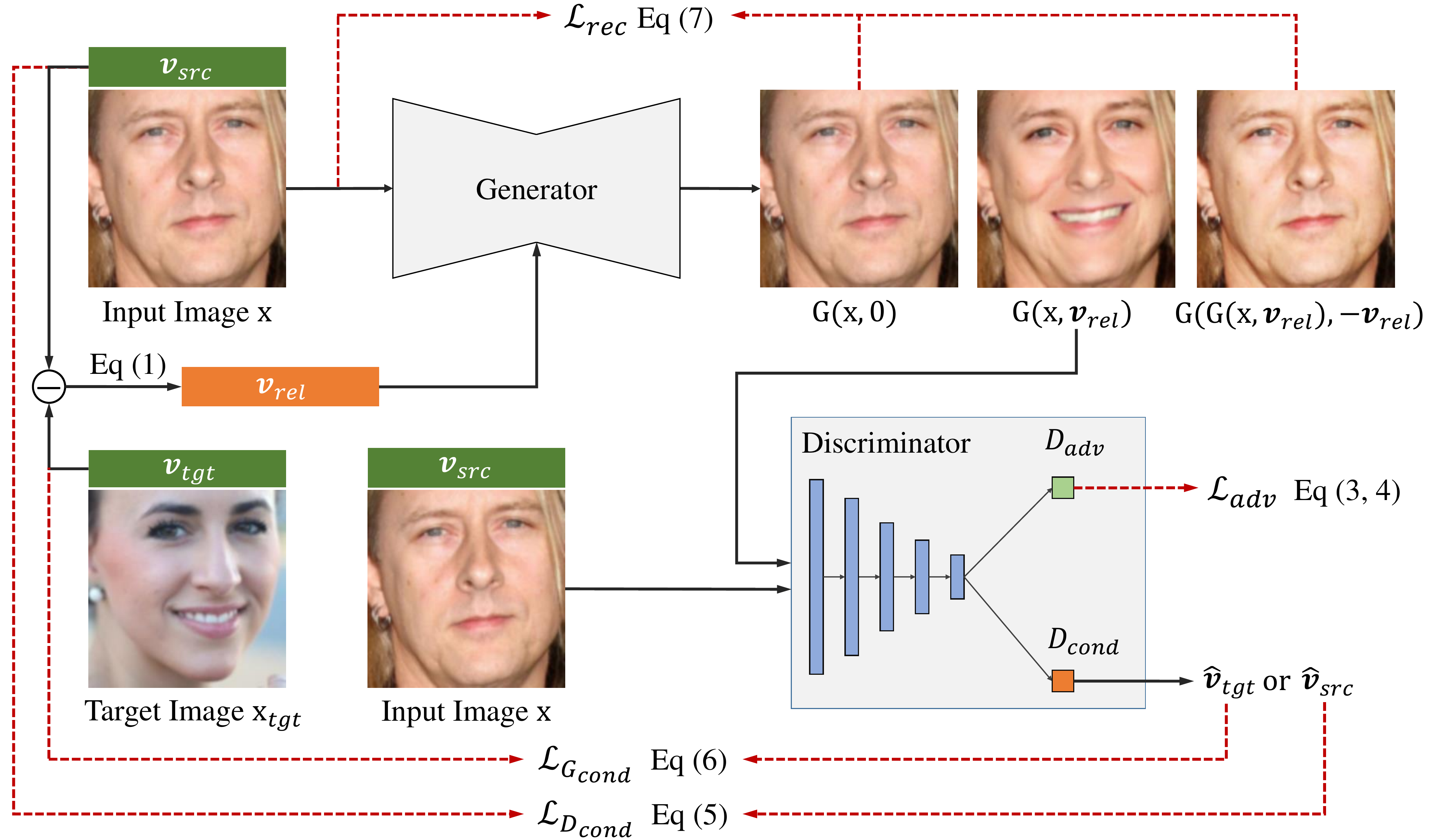}\\
   \caption{\textbf{An overview of proposed approach.} Our model consists of a single generator ($G$) and two discriminators ($D_{adv}$ and $D_{cond}$). (\emph{top}) $G$ conditions on an input image and relative action units to generate image with target expression. (\emph{bottom right}) $D_{adv}$ tries to distinguish between the input image $\mathbf{x}$ and generated image $G(\mathbf{x}, \mathbf{v}_{rel})$. Conditional discriminator $D_{cond}$ aims at evaluating generated image in condition fulfillment. }
\label{fig:framework_overview}
\end{figure}

For the purpose of higher image quality and better expression manipulation ability, we start from U-Net-based generator and analyze its limitations. Note that the features from encoder are directly concatenated with decoder features in U-Net structure. This often produces overlap artifacts when dealing with facial deformation. eIn this work, we resort to learn the model by simultaneously fusing and transforming image features at different spatial size. Particularly, we propose to introduce multi-resolution feature fusion mechanism and involve several multi-scale feature fusion (MSF) modules in basic U-Net architecture for image transformation. Taking relative AUs as conditional input, our MSF module adaptively fuse and modify both the features from the encoder and all lower resolution, and output fusion features with multi-resolution representation. The fusion features are further concatenated with decoder features for image decoding. Experimental results in Table~\ref{table:quantitative_network_based_metrics} and Fig~\ref{fig:training_curve} reveal the better expression manipulation ability and higher image quality brought by MSF mechanism and relative setting. Table~\ref{tabel:quantitative_expression_accuracy} and Fig~\ref{fig:single_au_comparison}, \ref{fig:expression_editing}, \ref{fig:hardsamples} demonstrate the superiority of our method compared to baseline model. An overview of our approach is provided in Fig.~\ref{fig:framework_overview}.


\section{Related Work}
\label{section_related_work}
\noindent
\textbf{Generative Adversarial Networks.} As one of the most promising unsupervised deep generative models, GANs~\cite{Goodfellow2014Generative} have achieved a series of impressive results. WGAN~\cite{arjovsky2017wassersteinGAN} stabilizes GAN training and alleviates model collapse problems by introducing Wasserstein distance. WGAN-GP~\cite{gulrajani2017wgangp} is suggested to improve WGAN training by enforcing gradient penalty. Conditional GAN~\cite{Mirza2014Conditional} generates images with desired properties under the constraint of extra conditional variables. Up to now, GANs have become one of the most prominent generative models in image synthesis~\cite{zhang2017stackgan,Karras_2019_CVPR_stylegan,peng2018variational}, super-resolution~\cite{wang2018esrgan,chu2018tecoGAN-videosr} and image-to-image translation~\cite{he2019attgan,park2019GauGAN}. 

\noindent
\textbf{Image-to-Image Translation.} Image-to-image translation can be treated as a cGAN that conditions on an image, aiming at learning an image mapping from one domain to another in supervised or unsupervised manner. Liu et al.~\cite{liu2017unsupervised} introduce a shared-latent space assumption and an unsupervised image-to-image translation framework based on Coupled GANs~\cite{liu2016coupled}. Pix2Pix~\cite{isola2017image-to-image} as well as~\cite{park2019GauGAN} is a supervised cGANs based approach which relies on an abundance of paired images. However, the absence of adequate paired data limits the performance of conditional GAN. To alleviate the dependency on paired images, Zhu et al.~\cite{zhu2017cyclegan} propose a cycle consistent framework for unpaired image-to-image translation. GANimation~\cite{Pumarola_ijcv2019} utilizes an encoder-decoder network to take images and entire action units as input to generate animated images but suffers from undesired artifacts in generated images.

\noindent
\textbf{Facial Expression Manipulation.} Facial expression manipulation is an interesting image-to-image translation problem, which has drawn prevalent attention recently. Some popular works tackle this task with multiple facial attributes editing~\cite{Choi_2018_CVPR_stargan,he2019attgan,liu2019stgan,powei_2019relgan}, modifying attribute categories such as to smiling, mouth open, mouth closed, adding beard, swapping gender and changing hair color, etc. However, these methods cannot simply generalize to an arbitrary human facial expression synthesizing tasks due to the limitations of discrete emotion categories(e.g., \emph{happy, neutral, surprised, contempt, anger, disgust, and sad}). Several studies, aiming at manipulating human facial expression from facial geometric representation~\cite{song2018geometry-guided,qiao2018geometry-contrastive}, conditioning on face fiducial points to synthesize animated faces but suffers from fine-grained details. Geng et al.~\cite{Geng_2019_CVPR_3DGuided} proposes a 3D parametric face guided model to manipulate the geometry of facial components, while requiring real existent target face images rather than a simple vector. 

\section{Methodology}
\label{section_ProposedMethod}
In this section, we present the components of our approach. We consider an input image as $\mathbf{x}$ with arbitrary facial expression. The expression is characterized by a one-dimensional AUs vector $\mathbf{v} = (v^1, ..., v^n)$, where each AU is normalized between 0 and 1 and $v^i$ indicates the intensity of the $i$-{th} AU. With the goal of translating $\mathbf{x}$ into a photo-realistic image, our generator takes relative AUs $\mathbf{v}_{rel}$ as condition to renders with target expression. In the following parts, relative action units, MSF module, network structure, and loss functions are presented. 

\subsection{Relative Action Units (AUs)}
\label{subsection_relativeAU}
Previous methods~\cite{Pumarola_ijcv2019} take both absolute target AUs vector $\mathbf{v}_{src}$ and source image $\mathbf{x}$ as input to the generator. However, this input setting is flawed in that the generator needs to estimate the real AUs of input image to determine whether to edit image. From an application perspective, we are required to provide a value that must be strictly equal to the corresponding AU in the source image (i.e., $v^i_{tgt}=v^i_{src}$, where $i=1,2,..., n$) even if we do not want to change it. Otherwise, the generator will probably introduce unintended modifications to editing results.

Compared to absolute AUs, relative AUs describe the desired change in selected action units. This is in accordance with the definition of action units~\cite{friesen1978facial} that indicates the activation state of facial muscles. Denote the source AUs and target AUs as $\mathbf{v}_{src}$ and $\mathbf{v}_{tgt}$. Therefore, the difference between target and source AUs can be defined as: 
\begin{equation}
\label{equ:relative_AU}
\mathbf{v}_{rel} \triangleq \mathbf{v}_{tgt} - \mathbf{v}_{src}
\end{equation}

Introducing relative AUs as input brings several benefits. First, the relative AUs represented by the difference between the source and target images are intuitive and user friendly. For example, if we only intend to suppress AU10 (Upper Lip Raiser), we could assign an arbitrary real negative value to $v_{10}$, while making the other values zero. Second, in comparison to entire target AUs, the values in $\mathbf{v}_{rel}$ are zero-centered and can provide more expressive information for guiding expression editing and stabilize the training process. Moreover, with relative AUs, the generator learns to edit and reconstruct facial parts with respect to non-zero and zero values, which alleviates the cost for action units preserving. In our experiments, $\mathbf{v}_{rel}$ with zero values hardly introduces artifacts and errors. 

Additionally, we propose to edit interpolated expressions $\mathbf{v}_{inter}$ among two different expressions $\mathbf{v}_{1}$ and $\mathbf{v}_{2}$. The interpolated AUs is denoted as Equation~\ref{equ:editing_strength}. 

\begin{equation}
\label{equ:editing_strength}
\mathbf{v}_{inter} = \mathbf{v}_{1} + \alpha (\mathbf{v}_{2} - \mathbf{v}_{1}) - \mathbf{v}_{src},\ 0 \le \alpha \le 1
\end{equation}

\subsection{Network Structure}
\label{subsection_network_structure}
As presented in Fig.~\ref{fig:fig_generator}(left), our generator is built on U-Net structure but replace several skip connections by our MSF modules in both high and low-resolution representation. The encoder consists of four convolutional layers with stride 2 for down-sampling, while the decoder is composed of four transposed convolutional layers with stride 2 for up-sampling. Furthermore, MSF module is applied as skip unit to fuse features from both higher and lower resolution in our generator. The kernel sizes are all $4\times4$ in down-sampling and up-sampling layers, while $5\times5$ in the rest convolutional layers. 

Our discriminator $D$ is the same as which in~\cite{Pumarola_ijcv2019}, which is trained to evaluate the generated images both in realism score and desired expression fulfillment. Two branches of the discriminator, namely $D_{adv}$ and $D_{cond}$, share a fully convolutional sub-network comprised of six convolutional layers with kernel size 4 and stride 2. On top of $D_{adv}$, we add a convolutional layer with kernel size 3, padding 1 and stride 1. For conditional critic $D_{cond}$, we add an auxiliary regression head to predict target AUs.

\subsection{Multi-Scale Feature Fusion}
\label{subsection_MSF_module}
\begin{figure}[!t]
\centering
   \includegraphics[width=1\linewidth]{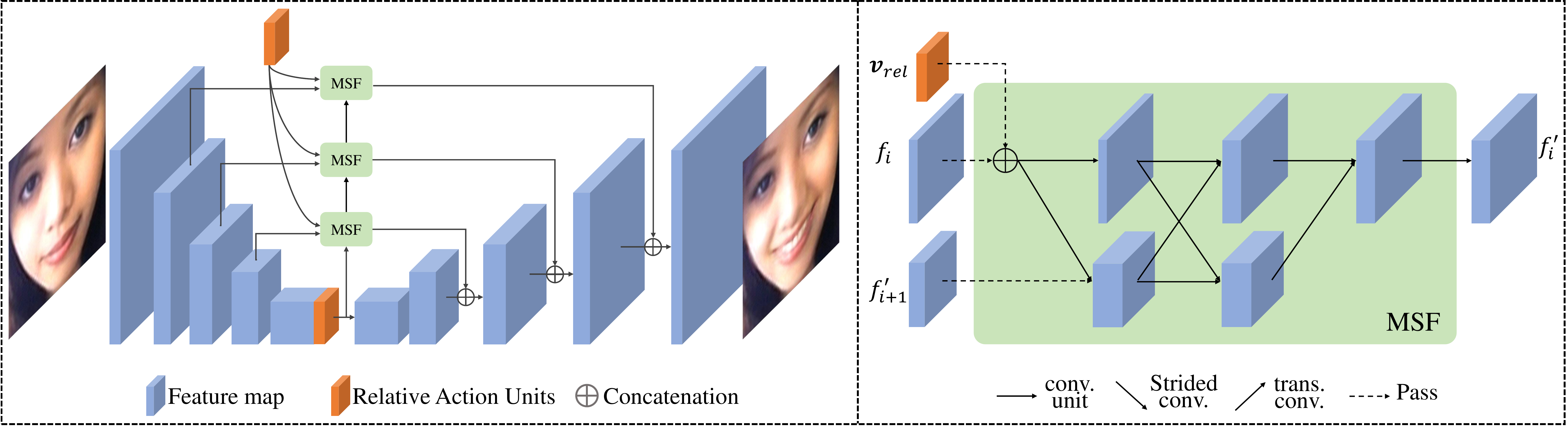}\\
   \caption{Left: the structure of our generator, incorporating several MSF modules which render the encoder features in different feature level. Right: details of the proposed MSF module. The bottom legend on the right figure: conv. = convolution, trans. = transposed. } 
\label{fig:fig_generator}
\end{figure}

Encoder-decoder architecture is insufficient to manipulate the image with high quality but U-Net based architectures support the rise of generating quality, according to ~\cite{liu2019stgan}. Taking these basics into consideration, we propose to modify the image features in different spatial resolution, simultaneously. To this end, we alter the structure in~\cite{sun2019deep_high_low} and then build a learnable sub-network, namely our multi-scale feature fusion (MSF) module, to manipulate features in multi-scale level. 
In Fig.~\ref{fig:fig_generator}(right), we show the overall architecture of multi-scale feature fusion module. 

Our MSF module is different from~\cite{sun2019deep_high_low} in two aspects. First, in MSF module, we fuse features from low-to-high, and the two kinds of conv streams noted above.  
In our approach, the MSF module takes the features across the encoder and the MSF modules as well as relative AUs as input and learns to manipulate image features at different spatial sizes. Second, we inject the condition at each MSF module. Such fusion mechanism helps MSF to learn the consistency of expressions of features with different resolutions, especially between encoder features and decoder features.

Without the loss of generality, we take the MSF module in $i$-th layer for example. Denote the input encoder features as $f_{i}$ from $i$-th layer of encoder, and fusion feature as $f'_{i+1}$ from the $i+1$-th MSF module. Firstly, the encoder features are concatenated with relative AU $\mathbf{v}_{rel}$ in depth-wise fashion. Then a convolutional unit and a down-sample layer are applied to acquire two feature maps in different spatial size. The down-sampled features are then concatenated with higher-level features $f'_{i+1}$ from $i+1$-th MSF module. One more parallel feature fusion unit is applied across high and low-resolution representation, and then formulated into the output $f'_{i}$. The fusion feature $f'_{i}$ will be one the input of decoder and $i-1$-th MSF module. In this way, our generator learns and transforms the image features collaboratively in a multi-scale manner.

\subsection{Loss Functions}
\label{subsection_loss_functions}
Denote the conditional generated image as  $\mathbf{x}^\prime = G(\mathbf{x}, \mathbf{v}_{rel})$, where input image $\mathbf{x}$ and relative attributes $\mathbf{v}_{rel}$ are considered as inputs of the generator. In the following, we will introduce the loss functions employed in our framework.

\noindent
\textbf{Adversarial Loss.} To synthesize photo-realistic images with GANs, we use the improved divergence criterion of standard GAN~\cite{Goodfellow2014Generative} proposed by WGAN-GP~\cite{gulrajani2017wgangp}. The adversarial loss can be written as:

\begin{equation}
\label{equ:adversarial_D}
\begin{split}
\max_{D_{adv}}\mathcal{L}_{D_{adv}} &= \mathbb{E}_{\mathbf{x}} D_{adv}(\mathbf{x}) - \mathbb{E}_{\mathbf{x}^\prime} D_{adv}(\mathbf{x}^\prime) - \lambda_{gp} \mathbb{E}_{\hat{\mathbf{x}}}[(\|\bigtriangledown_{\hat{\mathbf{x}}}D_{adv}(\hat{\mathbf{x}})\|_{2} - 1)^{2}]
\end{split}
\end{equation}
\begin{equation}
\label{equ:adversarial_G}
\max_{G} \mathcal{L}_{G_{adv}} = \mathbb{E}_{\mathbf{x}, \mathbf{v}_{rel}} D_{adv}(G(\mathbf{x}, \mathbf{v}_{rel}))
\end{equation}
where $\lambda_{gp}$ is a penalty coefficient and $\hat{\mathbf{x}}$ is randomly interpolated between $\mathbf{x}$ and generated image $\mathbf{x}^\prime$. The discriminator $D$ is unsupervised and aims to distinguish between real images and the generated fake images. The generator $G$  tries to generate images which look realistic as the real.

\noindent
\textbf{Conditional Fulfillment.} We require not only that the image synthesized by our model should look realistic, but also possess desired AUs. To this end, we adopt the core idea of conditional GANs~\cite{Mirza2014Conditional} and employ an action units regressor $D_{cond}$ which shares convolutional weights with $D_{adv}$, and define the following manipulation loss for training $D_{cond}$ and $G$:

\begin{equation}
\min_{D_{cond}}\mathcal{L}_{D_{cond}} = \mathbb{E}_{\mathbf{x}, \mathbf{v}_{src}}\|D_{cond}(\mathbf{x})-\mathbf{v}_{src}\|_2^{2}
\end{equation}
\begin{equation}
\min_{G}\mathcal{L}_{G_{cond}} = \mathbb{E}_{\mathbf{x}^\prime, \mathbf{v}_{tgt}}\|D_{cond}(\mathbf{x}^\prime) - \mathbf{v}_{tgt} \|_2^{2}
\end{equation}
where the AUs regression loss of real images $\mathbf{x}$ is used to optimize $D_{cond}$, thus $G$ can learn to generate images $\mathbf{x}^\prime$ which minimize the AUs regression loss $\mathcal{L}_{G_{cond}}$. 

\noindent
\textbf{Reconstruction Regularization.} Our generator $G$ is trained to generate an output image $G(\mathbf{x}, \mathbf{v}_{rel})$ which not only looks realistic but also possesses desired facial action units. However, there is no ground-truth supervision provided in the dataset for our model to modify facial components while preserving identity information. To this end, we add extra constraints to guarantee the faces in both input and output images are from the same person in appearance.

On one hand, we utilize a \emph{self-reconstruction} loss to enforce the generator to manipulate nothing when fed with zero-value relative AUs (i.e., $\mathbf{v}_{rel}=\mathbf{0}$). On the other hand, we adopt the concept of cycle consistency~\cite{zhu2017cyclegan} and formulate the \emph{cycle-reconstruction} loss which penalizes the difference between $G(G(\mathbf{x}, \mathbf{v}_{rel}), -\mathbf{v}_{rel})$ and the input source $\mathbf{x}$. Hence, thses two reconstruction losses can be written as:
\begin{equation}
\min_{G} \mathcal{L}_{rec} = \mathbb{E}_{\mathbf{x}}[\|\mathbf{x}-G(\mathbf{x}, \mathbf{0})\|_{1}] + \mathbb{E}_{\mathbf{x}, \mathbf{v}_{rel}}[\|G(G(\mathbf{x}, \mathbf{v}_{rel}), -\mathbf{v}_{rel})-\mathbf{x}\|_{1}]
\end{equation}
where $\mathbf{0}$ denotes a zero-padded vector with the same shape of $\mathbf{v}_{rel}$.

\noindent
\textbf{Total Variation Regularization.} To ensure smooth spatial transformation and naturalness of output images in RGB color space, we follow the prior work~\cite{johnson2016perceptual,Pumarola_ijcv2019} and perform a regularization $\mathcal{L}_{tv}$ over the synthesized fake samples $G(x, \mathbf{v}_{rel})$.

\noindent
\textbf{Model Objective.} Taking the above losses into account, we finally build our total loss functions for $D$ and $G$ by combining all previous partial losses, respectively, as:
\setlength{\parskip}{0em}
\begin{equation}
\label{equ:loss_D}
\min_{D} \mathcal{L}_{D} = -\mathcal{L}_{D_{adv}} + \lambda_{1}\mathcal{L}_{D_{cond}}
\end{equation}
\begin{equation}
\label{equ:loss_G}
\begin{split}
\min_{G} \mathcal{L}_{G} = &-\mathcal{L}_{G_{adv}} + \lambda_{2}\mathcal{L}_{G_{cond}} + \lambda_{3}\mathcal{L}_{rec} + \lambda_{4}\mathcal{L}_{tv}
\end{split}
\end{equation}
where $\lambda_{1}$, $\lambda_{2}$, $\lambda_{3}$, and $\lambda_{4}$ are tradeoff parameters that control the impact of each loss.

\section{Experiments}
\subsection{Implementation Details}
\label{subsection_implementation}

\noindent
\textbf{Dataset and Preprocessing.} We randomly choose a subset of 200,000 samples from AffectNet~\cite{mollahosseini2017affectnet} dataset. Besides, we remove some repeated images or cartoon faces in the validation set and take 3234 images as our testing samples to assess the training process. The images are centered cropped and resized to $128\times128$ by bicubic interpolation. All continuous AUs annotations are extracted by~\cite{baltrusaitis2018openface}. 

\noindent
\textbf{Baseline.} 
As the current state-of-the-art method, GANimation~\cite{Pumarola_ijcv2019}, outperforming plenty of representative facial expression synthesis models~\cite{Choi_2018_CVPR_stargan,zhu2017cyclegan,li2016diat,perarnau2016invertible}, is taken as our baseline model. For fair comparison, we use the code\footnote{https://github.com/albertpumarola/GANimation} released by the authors and train the model on AffectNet~\cite{mollahosseini2017affectnet} with default hyper-parameters. 

\noindent
\textbf{Experiment Settings.} We train the model by Adam~\cite{kingma2014adam} optimizer with settings of $\beta_1=0.5$, $\beta_2=0.999$ for 30 epochs at initial learning rate of $1\times 10^{-4}$, and then linearly decay the rate to $1\times 10^{-5}$ for fine-tuning. We perform every single optimization step of the generator with four optimization steps of the discriminator. The weight coefficients for Eqn.\ref{equ:loss_D} and \ref{equ:loss_G} are set to $\lambda_{1}=\lambda_{2}=150, \lambda_{3}=30, \lambda_{4}=5\times10^{-6}$. All experiments are conducted in PyTorch~\cite{paszke2017automatic} environment.

\subsection{Evaluation Metrics}
\label{subsection_evaluation_methods}
Evaluating a GAN model with respect to one criterion does not reliably reveal its convincing performance. In this work, we conduct model evaluation from two perspectives, which are network-based and human-based evaluation. Both methods measure the performance in three aspects, namely expression fulfillment, relative realism and identity preserving ability. 

\begin{figure}[!t]
\centering
   \includegraphics[width=1\linewidth]{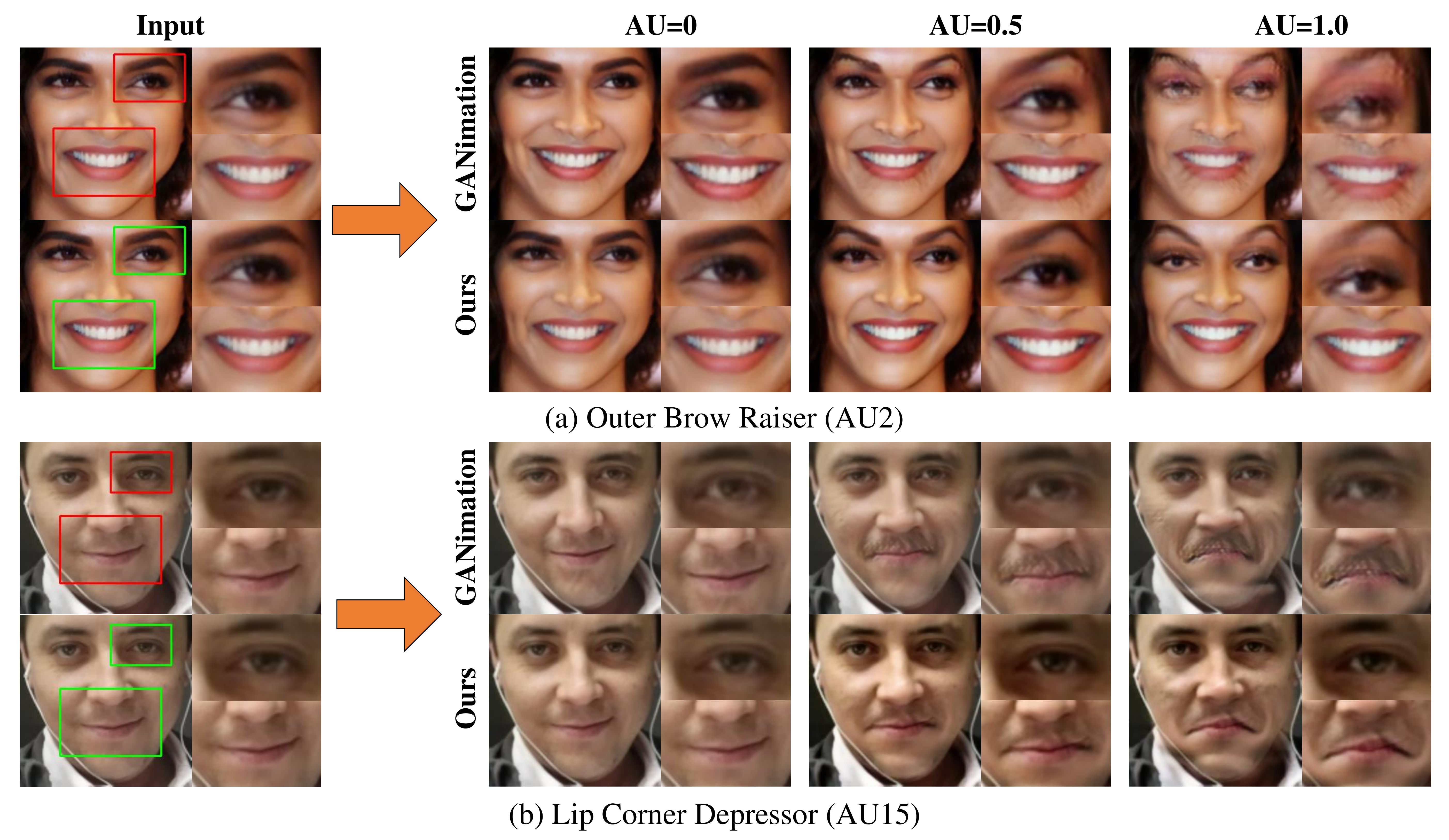}\\
   \caption{\textbf{Comparisons for single AU editing}. Each time, we manipulate the face with only one AU activated, leaving the rest part of face unchanged. The values upon images denote the relative AUs value in our test.} 
\label{fig:single_au_comparison}
\end{figure}

\noindent
\textbf{Network-based Metrics.} 
We evaluate 3234 images from AffectNet testing-set, each of which is transformed to 7 randomly selected expressions. Therefore, we get 22638 image pairs and then perform our quantitative evaluation.

\begin{enumerate}
   \item[$-$] \emph{Inception Score (IS)}. IS~\cite{salimans2016IS} utilizes an Inception network to extract image representation and calculates the KL divergence between the conditional distribution and marginal distribution. Although previous work~\cite{barratt2018note} has revealed the limitations of IS in intra-class images, it is still widely used to evaluate the model performance in image quality~\cite{Pumarola_ijcv2019,brock2018biggan}. Following the evaluating method in~\cite{Pumarola_ijcv2019}, we calculate IS of images synthesized by our approach and GANimation~\cite{Pumarola_ijcv2019}.

   \item[$-$] \emph{Average Content Distance (ACD).} ACD~\cite{Tulyakov2018MoCoGAN} measures $l_2$-distance between embedded features of the input and generated images. We employ a famous facial recognition network\footnote{https://github.com/ageitgey/face\_recognition}, as GANimation did in~\cite{Pumarola_ijcv2019}, to extract face code for each individual and calculate the distance for each expression editing result. The lower value indicates the better identity similarity between images before and after editing. 

   \item[$-$] \emph{Expression Distance (ED)}. 
   To consistently evaluate the ability of our model in expression editing, we reuse OpenFace2.0~\cite{baltrusaitis2018openface} to acquire the AUs of edited images, and calculate $l_2$-distance between the generated and target AUs (the lower, the better). Performing such objective evaluation is not trivial, as a categorized expression often related to two different AU intensity~\cite{friesen1978facial}. 
\end{enumerate}
\begin{figure}[!t]
\centering
   \includegraphics[width=1\linewidth]{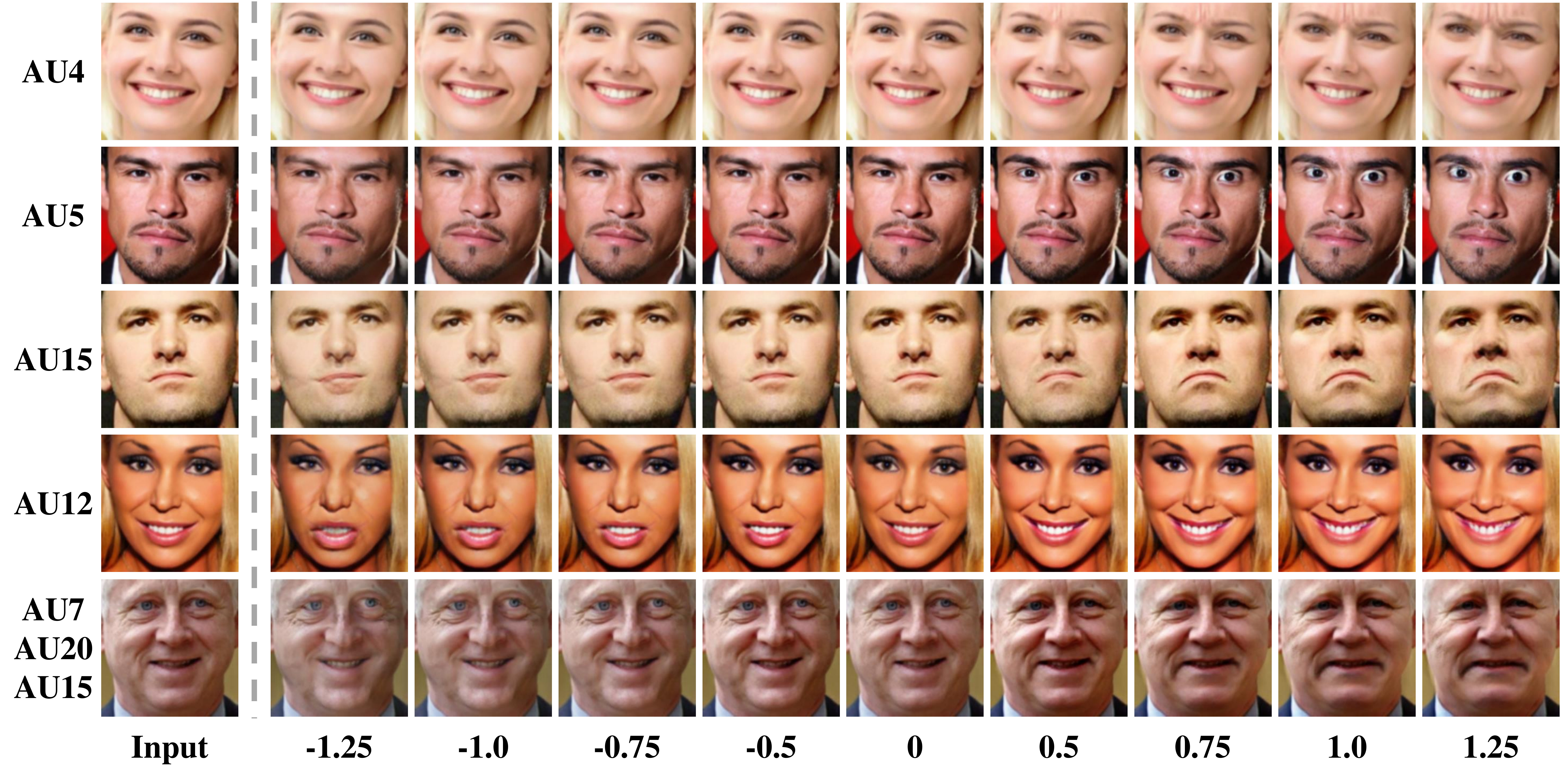}\\
   \caption{\textbf{Sample results in single/multiple AUs editing}. AU4: Brow Lowerer; AU5: Upper Lid Raiser; AU7: Lid Tightener; AU12: Lip Corner Puller; AU15: Lip Corner Depressor; AU20: Lip Stretcher. The legend below the images are relative AUs intensity. The higher (lower) AUs value means to strengthen (weaken) the corresponding facial action unit in input image. Please zoom in for better observation. }
\label{fig:single_au_editing}
\end{figure}

\noindent
\textbf{Human-based Metrics.} For each metric in human-based evaluation, we asked 20 volunteers to evaluate 100 pairs of images which are generated by baseline and our method. During the test, we randomly display the images and ensure that the users do not know which image is edited by our model. 

\begin{enumerate}
   \item[$-$] \emph{Relative realism}. In each comparison, we randomly select two images which are generated by GANimation and our model, respectively. The user is asked to pick the more realistic image they think. 

   \item[$-$] \emph{Identity preserving}. One more user study for identity similarity metric is conducted to verify if humans agree that the given two images are from the same person. The display order of synthesized images from GANimation or our model is random. 

   \item[$-$] \emph{Expression fulfillment}. Due to the complex distribution of human facial expressions, it is not very reasonable to classify expression into a specific category. To this end, we alleviate these limitations by asking the users to rate the similarity of two facial expressions instead of reporting the emotion labels. In every trail of human preference study, two images (one with target expression and the other one is edited by GANimation or our method) are displayed randomly. The users have to examine and rate the similarity of facial expressions in the two images. If the given two images are considered to be different in their opinion, the user is allowed to rate 0. When the user thinks the images are totally the same expression, 2 will be given. If the user is not sure about the similarity of two expressions or these expressions are partly the same (e.g., same AUs for mouth but different for eyebrows), 1 will be noted.

\end{enumerate}

\subsection{Qualitative Evaluation}
\label{subsection_qualitative_evaluation}
We first qualitatively compare our model with GANimation in edition of single or multiple AUs. Fig.~\ref{fig:single_au_comparison} shows two typical examples of AU2 (Outer Brow Raiser) and AU15 (Lip Corner Depressor). From sample results of (a) in Fig.~\ref{fig:single_au_comparison}, it can be observed that GANimation fails to focus on Outer Brow and wrinkles the mouth, yielding less satisfying results than ours. In sample results of sub-figure (b), our model produces more plausible and better-manipulated results, especially in regions around the lip corner.

Fig.~\ref{fig:single_au_editing} shows more results in single/multiple AUs editing. By adopting relative action units as conditional input, our model convincingly learns to edit a single or multiple AUs instead of entire action units of the input image. 

\begin{figure}[!htbp]
\centering
   \includegraphics[width=1\linewidth]{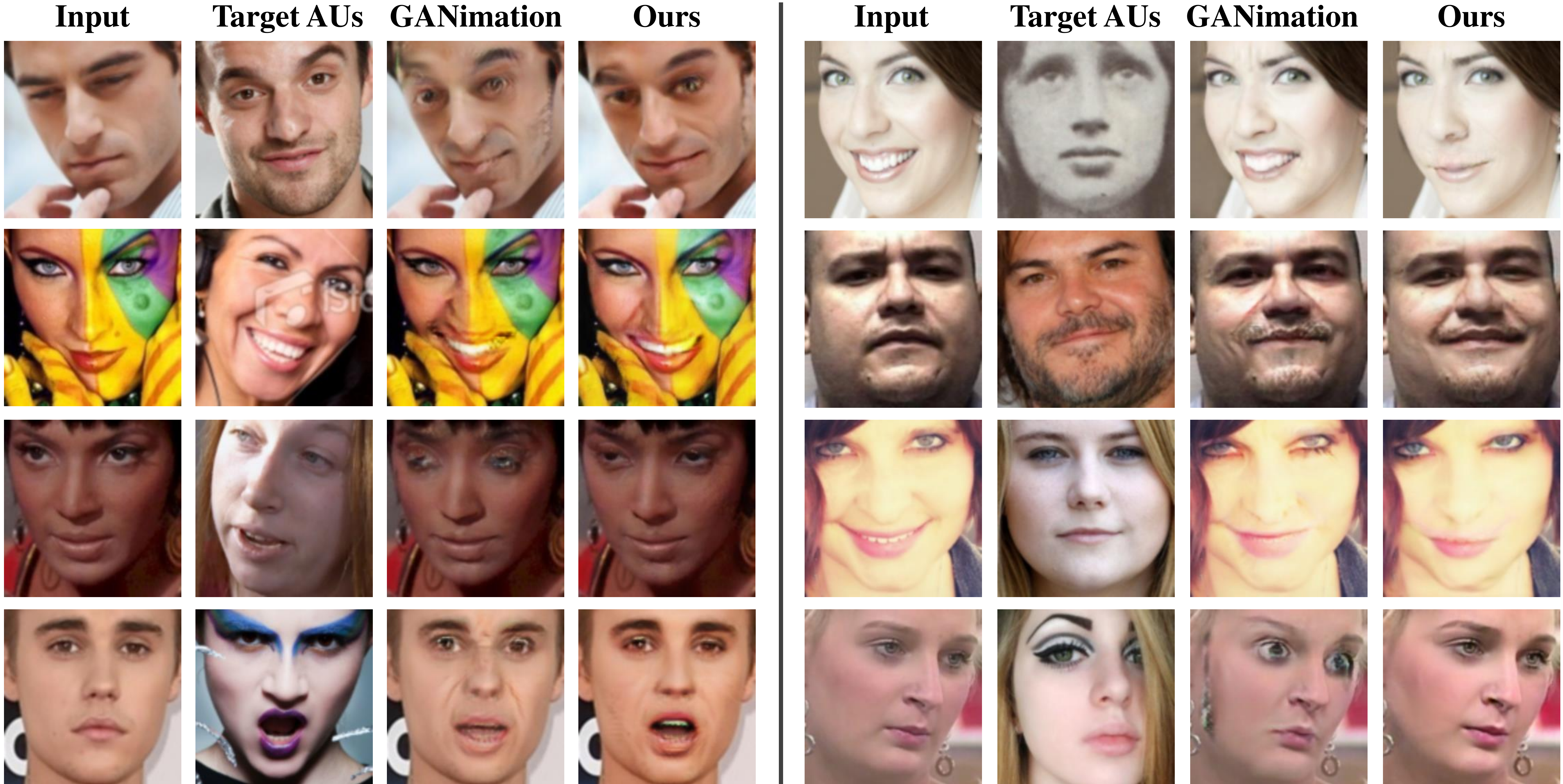}
   \caption{\textbf{Qualitative comparison}. Images are taken from AffectNet dataset.}
\label{fig:expression_editing}
\end{figure}

\begin{figure}[!ht]
\centering
   \includegraphics[width=1\linewidth]{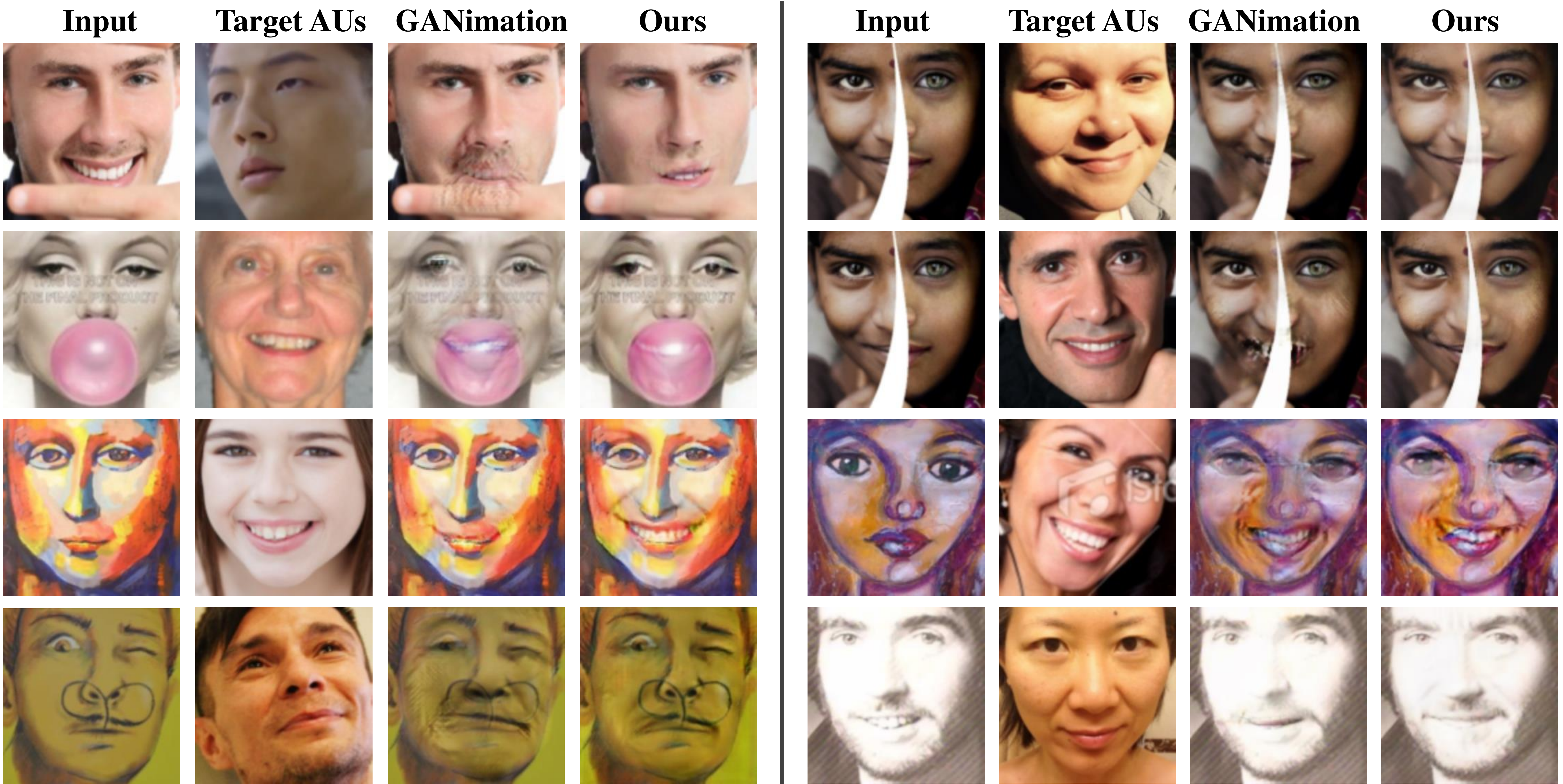}\\
   \caption{\textbf{Testing in difficult cases}. We compare our model to GANimation in several difficult cases, covering occlusions, paintings and drawings. }
\label{fig:hardsamples}
\end{figure}

\begin{figure}[!ht]
\centering
   \includegraphics[width=1\linewidth]{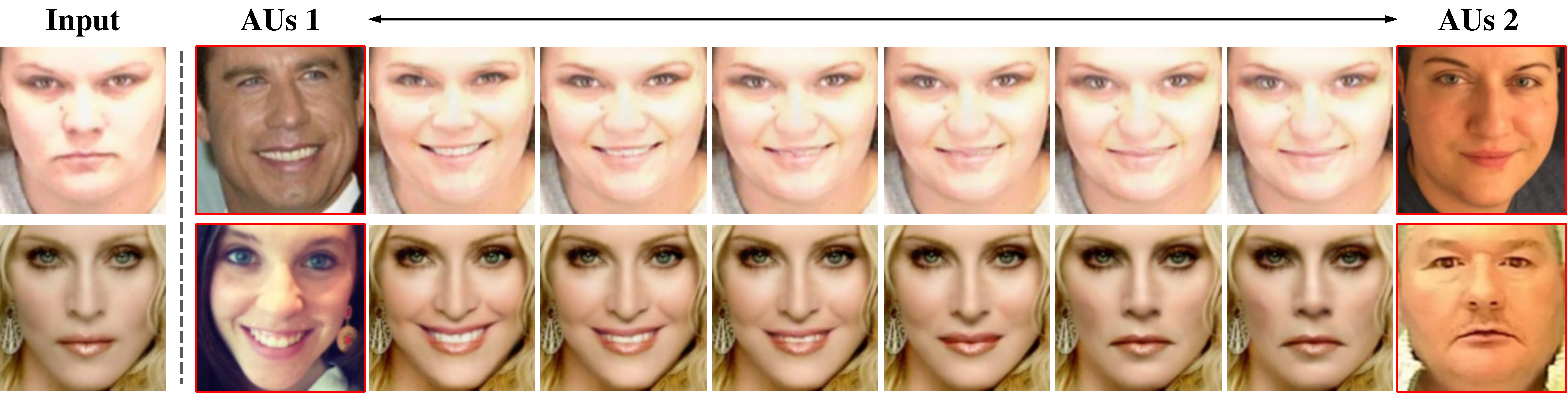}\\
   \caption{\textbf{Expression interpolation}. Example results of linear expression interpolation between two AUs vectors. }
\label{fig:expression_interpolation}
\end{figure}

We proceed to compare our model against GANimation. From the observation in Fig.~\ref{fig:expression_editing}, we can find that our model successfully transforms source image in accordance with desired AUs, with fewer artifacts and manipulation cues. While the baseline model is less likely to generate high-quality details or preserve the facial regions corresponding to unchanged AUs, especially for eyes and mouth.  

We next evaluate our network and discuss the model performance when dealing with extreme situations, which includes but not limited to image occlusions, portraits, drawings, and non-human faces. In Fig.~\ref{fig:hardsamples}, for instance, the first image shows occlusions created by a finger. To edit the expression for this kind of image, GANimation requires the entire set of AUs, including the activation status of Lip Corner and Chin, which imposes an extra burden on the user and brings an undesirable increase of visual artifacts. On the contrary, our method is able to edit expression without the need for source AUs. In the third and fourth row of Fig.~\ref{fig:hardsamples}, we present face editing examples from paintings and drawings, respectively. GANimation is either fails to efficiently manipulate input image with fully the same expression (the third row, left and the fourth row, right) or introduces unnatural artifacts and deformation (third row, right and fourth row, left). We can easily find the improvements of our method when compared to GANimation, although GANimation achieves plausible results on these images.

To better understand the benefits of continuous editing, we exploit AUs interpolation between different expressions and present results in Fig.~\ref{fig:expression_interpolation}. The plausible results verify the continuity in the action units space and demonstrate the generalization performance of our model. 


\subsection{Quantitative Evaluation}
\label{subsection:quantitative_evaluation}
Here we will conduct quantitative evaluations to verify the qualitative comparisons above. As described in Sec.~\ref{subsection_evaluation_methods}, we resort to three alternative measures for quantitative evaluation of our method. First, we calculate metrics of IS, ACD and ED for both GANimation and the proposed approach. The comparison results are given in Table~\ref{table:quantitative_network_based_metrics}. It can be observed that our approach consistently achieves competitive results against GANimation for IS and ED. Our generator without MSF module attains the lowest score on ACD but the highest score in ED. This is reasonable because the accuracy of a facial recognition network inevitably suffers from expression variation. 

\begin{table}[!ht]
\small
\centering
\caption{\textbf{Network-based evaluation.} Better results are in bold.}
\label{table:quantitative_network_based_metrics}
\begin{tabular}{|p{4cm}<{\centering}|p{2.4cm}<{\centering}|p{1.6cm}<{\centering}|p{1.6cm}<{\centering}|}
\hline
Method & IS $\uparrow$ & ACD $\downarrow$ & ED $\downarrow$\\
\hline
Real Images & 3.024$\pm$0.157 & - & - \\
\hline
GANimation & 2.861$\pm$0.054 & 0.395 & 0.313 \\
GANimation w/ $v_{rel}$ & 2.901$\pm$0.043 & 0.352 & 0.661 \\
\hline
Ours, ${k_{MSF}=0}$ & 2.809$\pm$0.058 & \textbf{0.335} & 0.636 \\
Ours, ${k_{MSF}=1}$ & 2.864$\pm$0.042 & 0.349 & 0.609 \\
Ours, ${k_{MSF}=2}$ & 2.899$\pm$0.038 & 0.345 & 0.422 \\
Ours & \textbf{2.940$\pm$0.039} & 0.375 & \textbf{0.275} \\
Ours w/o $v_{rel}$ & 2.808$\pm$0.050 & 0.426 & 0.290 \\
\hline
\end{tabular}
\end{table}
\begin{table}
\footnotesize
\centering
\caption{\textbf{Human-based evaluation.} We presnet the proportion of user subjective evaluation on edited expression fulfillment and human preference. Better results are in bold.}
\label{tabel:quantitative_expression_accuracy}
\begin{tabular}{|p{2.5cm}<{\centering}|p{1.2cm}<{\centering}|p{1.2cm}<{\centering}|p{1.2cm}<{\centering}|p{1.6cm}<{\centering}|p{1.6cm}<{\centering}|}

	\hline
    \multirow{2}{*}{Method} & \multicolumn{3}{c|}{Expression Similarity} & \multicolumn{2}{c|}{Human Perference} \\
    \cline{2-6}
     & 0 $\downarrow$ & 1 $\uparrow$ & 2 $\uparrow$ & Realism $\uparrow$ & Identity $\uparrow$ \\
    \hline
    GANimation & 25.04 & \textbf{43.25} & 31.71 & 34.43 & \textbf{90.59} \\
    \hline
    Ours & \textbf{17.66} & 35.22 & \textbf{47.12} & \textbf{65.57} & 90.56 \\
    \hline
\end{tabular}
\end{table}

\begin{table}[!htbp]
\footnotesize
\centering
\caption{\textbf{Reconstruction comparison}. We measure the reconstruction error using L1 distance (lower is better), PSNR and SSIM~\cite{wang2004image} (higher is better). }
\label{table:reconstruction_comparison}
\begin{tabular}{|p{3.5cm}<{\centering}|p{1.4cm}<{\centering}|p{1.4cm}<{\centering}|p{1.4cm}<{\centering}|}
\hline
Method & L1 $\downarrow$ & PSNR $\uparrow$ & SSIM $\uparrow$ \\
\hline
GANimation  & 0.049 & 23.76 & 0.901 \\
GANimation w/ $v_{rel}$ & 0.022 & 29.11 & 0.954 \\
\hline
Ours w/o $v_{rel}$ & 0.025 & 28.83 & 0.972 \\
Ours & 0.018 & 31.89 & 0.986 \\
\hline
\end{tabular}
\end{table}

Table~\ref{tabel:quantitative_expression_accuracy}, as a supplement to metric ED, offers a human-based evaluation on expression editing ability. Benefiting from MSF modules which serve as skip connections from encoder to decoder, our approach outperforms GANimation by a large margin. Nearly a quarter of test samples transformed by GANimation are considered failures. The proposed model is slightly favorable to the baseline in terms of identity preservation and our model performs better in image realism score, according to human preference results in Table~\ref{tabel:quantitative_expression_accuracy}(right part).

\subsection{Ablation Study}
\label{subsection_ablationstudy}
In this section, we exploit the importance of each component within the proposed method. To begin with, we investigate the improvement brought by relative AUs. We compare our model with baseline model in action units preserving from reconstruction perspective. To perform facial image reconstruction, we respectively apply GANimation by taking source AUs as absolute condition, and apply our model by taking a zero-valued vector as relative condition. We present results of L1 norm, PSNR, and SSIM~\cite{wang2004image} between input and generated images in Table~\ref{table:reconstruction_comparison}. From the second and third row, it can be seen that GANimation trained with relative AUs is slightly better than our approach without using relative AUs. When trained with our full approach (fourth row), we achieve the best reconstruction results. 

\begin{figure}[!ht]
\centering
   \includegraphics[width=1\linewidth]{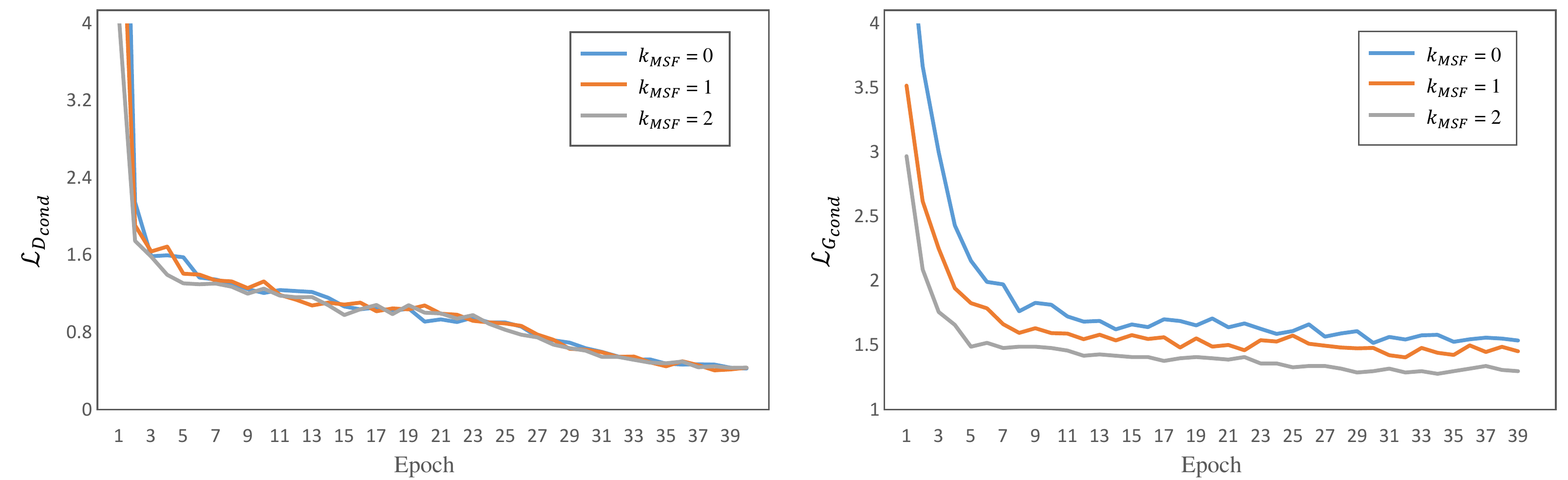}
    \caption{\textbf{Conditional loss convergence.} Left(right) figure: the learning curves of condition loss in discriminator(generator). We use the same discriminator during ablation study. }
\label{fig:training_curve}
\end{figure}

We next examine the importance of MSF module based on IS/ACD/ED metrics. Note that our model is built on U-Net, we carefully replace the skip connection with our MSF module and gradually train these generators separately. Quantitative comparison results are shown in Table~\ref{table:quantitative_network_based_metrics}. The first case is our model without MSF module (fourth row), which reduces to U-Net architecture. U-Net-based model acquires the best ACD result and the worst expression distance, which implies inefficient performance in expression editing. A conclusion can be drawn from the comparison results that a model has a greater potential to attain lower ACD if the ED gets higher. One proper explanation is that the expression editing intensity inevitably change the face features for facial recognition network. Fig.~\ref{fig:training_curve} shows the loss optimization process in our experiments. As can be found, the trend of loss curves are almost the same during the period of training discriminator (left figure). From the right figure, we can find that the generator that has two MSF modules converges faster than those with less MSF modules, which implies the definite improvements are brought by our MSF mechanism.

\section{Conclusion}
In this study, we propose a novel approach by incorporating multi-scale fusion mechanism in U-Net based architecture for arbitrary facial expression editing. As a simple but competitive method, relative condition setting is proved to improve our model performance by a large margin, especially for action units preserving, reconstruction quality and identity preserving. We achieve better experimental results in visual quality, manipulation ability, and human preference compared to state-of-the-art methods. 

\noindent
\textbf{Acknowledgements.}
This work was supported by NSFC (61671296, U1611461), National Key R$\&$D Project of China(2019YFB1802701), MoE-China Mobile Research Fund Project(MCM20180702) and the Shanghai Key Laboratory of Digital Media Processing and Transmissions.

%
%
\bibliographystyle{splncs04}
\bibliography{egbib}

\begin{thebibliography}{10}
\providecommand{\url}[1]{\texttt{#1}}
\providecommand{\urlprefix}{URL }
\providecommand{\doi}[1]{https://doi.org/#1}

\bibitem{arjovsky2017wassersteinGAN}
Arjovsky, M., Chintala, S., Bottou, L.: Wasserstein generative adversarial
  networks. In: Proceedings of the 34th International Conference on Machine
  Learning. pp. 214--223 (2017)

\bibitem{baltrusaitis2018openface}
Baltrusaitis, T., Zadeh, A., Lim, Y.C., Morency, L.P.: Openface 2.0: Facial
  behavior analysis toolkit. In: 2018 13th IEEE International Conference on
  Automatic Face \& Gesture Recognition (FG 2018). pp. 59--66. IEEE (2018)

\bibitem{barratt2018note}
Barratt, S., Sharma, R.: A note on the inception score. arXiv preprint
  arXiv:1801.01973  (2018)

\bibitem{brock2018biggan}
Brock, A., Donahue, J., Simonyan, K.: Large scale gan training for high
  fidelity natural image synthesis. In: International Conference on Learning
  Representation (2019)

\bibitem{Choi_2018_CVPR_stargan}
Choi, Y., Choi, M., Kim, M., Ha, J.W., Kim, S., Choo, J.: Stargan: Unified
  generative adversarial networks for multi-domain image-to-image translation.
  In: Proceedings of the IEEE Conference on Computer Vision and Pattern
  Recognition. pp. 8789--8797 (2018)

\bibitem{chu2018tecoGAN-videosr}
Chu, M., Xie, Y., Leal-Taix{\'e}, L., Thuerey, N.: Temporally coherent gans for
  video super-resolution (tecogan). ACM Transactions on Graphics (TOG)
  \textbf{39}(4),  75--1 (2020)

\bibitem{friesen1978facial}
Friesen, E., Ekman, P.: Facial action coding system: a technique for the
  measurement of facial movement. Palo Alto  \textbf{3} (1978)

\bibitem{Geng_2019_CVPR_3DGuided}
Geng, Z., Cao, C., Tulyakov, S.: 3d guided fine-grained face manipulation. In:
  Proceedings of the IEEE Conference on Computer Vision and Pattern
  Recognition. pp. 9821--9830 (2019)

\bibitem{Goodfellow2014Generative}
Goodfellow, I., Pouget-Abadie, J., Mirza, M., Xu, B., Warde-Farley, D., Ozair,
  S., Courville, A., Bengio, Y.: Generative adversarial nets. In: Advances in
  Neural Information Processing Systems. pp. 2672--2680 (2014)

\bibitem{gulrajani2017wgangp}
Gulrajani, I., Ahmed, F., Arjovsky, M., Dumoulin, V., Courville, A.C.: Improved
  training of wasserstein gans. In: Advances in Neural Information Processing
  Systems. pp. 5767--5777 (2017)

\bibitem{he2019attgan}
He, Z., Zuo, W., Kan, M., Shan, S., Chen, X.: Attgan: Facial attribute editing
  by only changing what you want. IEEE Transactions on Image Processing
  \textbf{28}(11),  5464--5478 (2019)

\bibitem{isola2017image-to-image}
Isola, P., Zhu, J.Y., Zhou, T., Efros, A.A.: Image-to-image translation with
  conditional adversarial networks. In: Proceedings of the IEEE Conference on
  Computer Vision and Pattern Recognition. pp. 1125--1134 (2017)

\bibitem{johnson2016perceptual}
Johnson, J., Alahi, A., Fei-Fei, L.: Perceptual losses for real-time style
  transfer and super-resolution. In: European Conference on Computer Vision.
  pp. 694--711. Springer (2016)

\bibitem{Karras_2019_CVPR_stylegan}
Karras, T., Laine, S., Aila, T.: A style-based generator architecture for
  generative adversarial networks. In: Proceedings of the IEEE Conference on
  Computer Vision and Pattern Recognition. pp. 4401--4410 (2019)

\bibitem{kingma2014adam}
Kingma, D.P., Ba, J.: Adam: A method for stochastic optimization. arXiv
  preprint arXiv:1412.6980  (2014)

\bibitem{li2016diat}
Li, M., Zuo, W., Zhang, D.: Deep identity-aware transfer of facial attributes.
  arXiv preprint arXiv:1610.05586  (2016)

\bibitem{liu2019stgan}
Liu, M., Ding, Y., Xia, M., Liu, X., Ding, E., Zuo, W., Wen, S.: Stgan: A
  unified selective transfer network for arbitrary image attribute editing. In:
  Proceedings of the IEEE Conference on Computer Vision and Pattern
  Recognition. pp. 3673--3682 (2019)

\bibitem{liu2017unsupervised}
Liu, M.Y., Breuel, T., Kautz, J.: Unsupervised image-to-image translation
  networks. In: Advances in Neural Information Processing Systems. pp. 700--708
  (2017)

\bibitem{liu2016coupled}
Liu, M.Y., Tuzel, O.: Coupled generative adversarial networks. In: Advances in
  Neural Information Processing Systems. pp. 469--477 (2016)

\bibitem{Mirza2014Conditional}
Mirza, M., Osindero, S.: Conditional generative adversarial nets  (2014)

\bibitem{mollahosseini2017affectnet}
Mollahosseini, A., Hasani, B., Mahoor, M.H.: Affectnet: A database for facial
  expression, valence, and arousal computing in the wild. IEEE Transactions on
  Affective Computing  \textbf{10}(1),  18--31 (2017)

\bibitem{park2019GauGAN}
Park, T., Liu, M.Y., Wang, T.C., Zhu, J.Y.: Semantic image synthesis with
  spatially-adaptive normalization. In: Proceedings of the IEEE Conference on
  Computer Vision and Pattern Recognition. pp. 2337--2346 (2019)

\bibitem{paszke2017automatic}
Paszke, A., Gross, S., Chintala, S., Chanan, G., Yang, E., DeVito, Z., Lin, Z.,
  Desmaison, A., Antiga, L., Lerer, A.: Automatic differentiation in pytorch
  (2017)

\bibitem{peng2018variational}
Peng, X.B., Kanazawa, A., Toyer, S., Abbeel, P., Levine, S.: Variational
  discriminator bottleneck: Improving imitation learning, inverse rl, and gans
  by constraining information flow. arXiv preprint arXiv:1810.00821  (2018)

\bibitem{perarnau2016invertible}
Perarnau, G., Van De~Weijer, J., Raducanu, B., {\'A}lvarez, J.M.: Invertible
  conditional gans for image editing. arXiv preprint arXiv:1611.06355  (2016)

\bibitem{Pumarola_ijcv2019}
Pumarola, A., Agudo, A., Martinez, A., Sanfeliu, A., Moreno-Noguer, F.:
  Ganimation: One-shot anatomically consistent facial animation  (2019)

\bibitem{qiao2018geometry-contrastive}
Qiao, F., Yao, N., Jiao, Z., Li, Z., Chen, H., Wang, H.: Geometry-contrastive
  gan for facial expression transfer. arXiv preprint arXiv:1802.01822  (2018)

\bibitem{salimans2016IS}
Salimans, T., Goodfellow, I., Zaremba, W., Cheung, V., Radford, A., Chen, X.:
  Improved techniques for training gans. In: Advances in Neural Information
  Processing Systems. pp. 2234--2242 (2016)

\bibitem{song2018geometry-guided}
Song, L., Lu, Z., He, R., Sun, Z., Tan, T.: Geometry guided adversarial facial
  expression synthesis. In: 2018 ACM Multimedia Conference on Multimedia
  Conference. pp. 627--635. ACM (2018)

\bibitem{sun2019deep_high_low}
Sun, K., Xiao, B., Liu, D., Wang, J.: Deep high-resolution representation
  learning for human pose estimation. In: Proceedings of the IEEE Conference on
  Computer Vision and Pattern Recognition. pp. 5693--5703 (2019)

\bibitem{Tulyakov2018MoCoGAN}
Tulyakov, S., Liu, M.Y., Yang, X., Kautz, J.: {MoCoGAN}: Decomposing motion and
  content for video generation. In: IEEE Conference on Computer Vision and
  Pattern Recognition. pp. 1526--1535 (2018)

\bibitem{wang2018esrgan}
Wang, X., Yu, K., Wu, S., Gu, J., Liu, Y., Dong, C., Qiao, Y., Change~Loy, C.:
  Esrgan: Enhanced super-resolution generative adversarial networks. In:
  Proceedings of the European Conference on Computer Vision (ECCV). pp.~0--0
  (2018)

\bibitem{wang2004image}
Wang, Z., Bovik, A.C., Sheikh, H.R., Simoncelli, E.P.: Image quality
  assessment: from error visibility to structural similarity. IEEE Transactions
  on Image Processing  \textbf{13}(4),  600--612 (2004)

\bibitem{powei_2019relgan}
Wu, P.W., Lin, Y.J., Chang, C.H., Chang, E.Y., Liao, S.W.: Relgan: Multi-domain
  image-to-image translation via relative attributes. In: Proceedings of the
  IEEE International Conference on Computer Vision. pp. 5914--5922 (2019)

\bibitem{Zhang2018sagenerative}
Zhang, G., Kan, M., Shan, S., Chen, X.: Generative adversarial network with
  spatial attention for face attribute editing. In: Proceedings of the European
  Conference on Computer Vision. pp. 417--432 (2018)

\bibitem{zhang2017stackgan}
Zhang, H., Xu, T., Li, H., Zhang, S., Wang, X., Huang, X., Metaxas, D.N.:
  Stackgan: Text to photo-realistic image synthesis with stacked generative
  adversarial networks. In: Proceedings of the IEEE International Conference on
  Computer Vision. pp. 5907--5915 (2017)

\bibitem{zhu2017cyclegan}
Zhu, J.Y., Park, T., Isola, P., Efros, A.A.: Unpaired image-to-image
  translation using cycle-consistent adversarial networks. In: Proceedings of
  the IEEE International Conference on Computer Vision. pp. 2223--2232 (2017)

\end{thebibliography}
\end{document}